\documentclass{article}

\usepackage[nonatbib,preprint]{neurips_2023}
\usepackage{enumitem}
\usepackage{xcolor}
\usepackage{amsmath, bm, mathtools}
\usepackage{multirow, xspace}

\usepackage[many]{tcolorbox}

\newcommand{\F}{Fig.}
\newcommand{\T}{Table}
\renewcommand{\S}{Sec.}

\newcommand{\ignore}[1]{}

\newcommand{\tool}{EAC\xspace}

\usepackage{pifont}

\usepackage[utf8]{inputenc} 
\usepackage[T1]{fontenc}    
\usepackage{hyperref}       
\usepackage{url}            
\usepackage{booktabs}       
\usepackage{amsfonts}       
\usepackage{nicefrac}       
\usepackage{microtype}      

\title{Explain Any Concept: Segment Anything Meets Concept-Based Explanation}

\author{Ao Sun, Pingchuan Ma, Yuanyuan Yuan, and Shuai
  Wang\\
  The Hong Kong University of Science and Technology
  \\
  \tt aosun3@illinois.edu \{pmaab, yyuanaq, shuaiw\}@cse.ust.hk 
}

\begin{document}

\maketitle

\begin{abstract}
    EXplainable AI (XAI) is an essential topic to improve human understanding of
    deep neural networks (DNNs) given their black-box internals. For computer
    vision tasks, mainstream pixel-based XAI methods explain DNN decisions by
    identifying important pixels, and emerging concept-based XAI explore forming explanations
    with concepts (e.g., a head in an image). However, pixels are generally hard
    to interpret and sensitive to the imprecision of XAI methods, whereas
    ``concepts'' in prior works require human annotation or are limited to
    pre-defined concept sets.
    %
    On the other hand, driven by large-scale pre-training, Segment Anything Model
    (SAM) has been demonstrated as a powerful and promotable framework for performing
    precise and comprehensive instance segmentation, enabling automatic preparation
    of concept sets from a given image.
    This paper for the first time explores using SAM to augment
    concept-based XAI. We offer an effective and flexible 
    concept-based explanation method, namely Explain Any Concept (EAC), which 
    explains DNN decisions with any concept. 
    While SAM is highly effective and offers an ``out-of-the-box'' instance
    segmentation, it is costly when being integrated into de facto XAI
    pipelines. We thus propose a lightweight per-input equivalent (PIE) scheme,
    enabling efficient explanation with a surrogate model. Our evaluation over
    two popular datasets (ImageNet and COCO) illustrate the highly encouraging
    performance of \tool\ over commonly-used XAI methods.
\end{abstract}

\section{Introduction}

In recent years, Deep Neural Networks (DNNs) have exhibited exceptional
performance in a variety of computer vision (CV) tasks such as image
classification~\cite{he2016deep}, object detection~\cite{ren2015faster}, and
semantic segmentation~\cite{long2015fully}. However, due to the ``black-box''
nature of these complex models, their use in security-sensitive applications
where interpretability is critical is still limited. As a result, there is a
growing demand for increased transparency and comprehensibility in the
decision-making processes of DNNs. To address this issue, Explainable AI
(XAI)~\cite{adadi2018peeking} has emerged with the purpose of providing
explanations for DNNs' predictions.

For CV tasks, conventional XAI works primarily focus on proposing
and enhancing pixel-level interpretation, which explains the model prediction by
identifying important pixels. Despite the strides made in XAI, these techniques
often involve trade-offs between three key desiderata among a bunch of criteria:
\textit{faithfulness}, \textit{understandability}, and \textit{efficiency}~\cite{li2021instance,
hsiao2021roadmap, herm2023impact}. Faithfulness ensures that the generated
explanation aligns with the internal decision-making process of the DNN,
understandability ensures that the explanations are human-comprehensible, and
efficiency guarantees that the explanations are generated with a reasonable
computational cost.

From the perspective of explanation forms, existing methods often provide pixel
or superpixel-level
explanations~\cite{simonyan2013deep,fong2017interpretable,zhou2016learning,selvaraju2017grad,ribeiro2016should},
which are constantly hard to interpret (i.e., low understandability) and
sensitive to the potential imprecision of XAI techniques (low faithfulness).
Some recent works aim to explain DNN predictions with concepts (e.g., a head
rather than pixels in the head) in images. However, they either require human
annotation or are limited specific concept discovery methods. 

From the XAI methodology perspective, Shapley
value~\cite{shapley1953value}-based explanation has become the mainstream, given
its well-established guarantee in theory. However, the inherent complexity of
Shapley value calculations and the target model makes it highly costly and
time-consuming (\textit{low efficiency}). To reduce the high overhead, existing
methods rely on Monte Carlo sampling~\cite{song2016shapley}, model-specific
approximations~\cite{lundberg2020local,ancona2019explaining}, (smaller)
surrogate models~\cite{lundberg2017unified,covert2021improving}, or a
combination of the above to approximate the Shapley value. With the growth of
model size and complexity, surrogate models are widely employed, even though it
may suffer from low faithfulness owing to the discrepancy between the surrogate
model and the target model under consideration.

\noindent \textbf{Our Solution.}~Driven by large-scale pre-training, Segment
Anything Model (SAM)~\cite{kirillov2023segment} has been demonstrated as a
powerful and promotable framework for performing precise and comprehensive
instance segmentation, enabling automatic extraction of a concept set from an
given image.
Hence, this paper for the first time explores using SAM as a concept discovery
method to augment concept-based XAI. We advocate to line up SAM and XAI, such
that SAM's instance segmentation delivers high accurate and human-understandable
concept set from arbitrary images, which in turn facilitates the XAI task with
high faithfulness and understandability.

Nevertheless, while SAM is effective and offers an ``out-of-the-box'' solution,
computing the Shapley value is still expensive. Thus, to achieve high efficiency
(our third desiderata), besides standard Monte Carlo sampling method to reduce
overheads, we propose a lightweight \textit{per-input equivalent (PIE) scheme}
to approximate the target model with a low-cost surrogate model. Our PIE
scheme allows the surrogate model to share some carefully-chosen
parameters with the target model, which effectively close the discrepancy
between the two models.



Our evaluation over two popular datasets (ImageNet and COCO) illustrate the
highly encouraging and superior accuracy of EAC over popular pixel-level and
superpixel-based XAI methods. Moreover, we also demonstrate the high
interptertability of EAC with a carefully-designed user study. As confirmed by
human experts, EAC offers high interpretability, largely outperforming de facto
XAI methods. We also justify the effectiveness of our technical pipeline with
ablation studies, and discuss potential extension and future work. In summary,
this paper makes the following contributions:

\begin{itemize}
  \item We for the first time advocate the usage of SAM as a concept discovery
    method to facilitate concept-based XAI with high faithfulness and
    understandability.

  \item We propose a general and flexible concept-based explanation pipeline,
  namely Explain Any Concept (\tool), which can explain the model prediction
  with any concept. We introduce a set of design considerations and
  optimizations to make \tool practically efficient while maintaining high
  faithfulness and understandability.

  \item We conduct extensive experiments and human studies to demonstrate the
  effectiveness of EAC on diverse settings. We also illustrate the
  generalizability of \tool\ and its security benefits. 
  
\end{itemize}

\noindent \textbf{Open Source.}~We will publicly release and maintain \tool\
after its official publishing to benefit the community and follow-up usage.
\section{Background and Related Works}
\label{sec:bg}

\tool\ belongs to local XAI which offers model-agnostic explanations for DNN
decisions over each image input. Below, we introduce prior works from both XAI
methodology and input data perspective.

\noindent \textbf{XAI Methodology.}~From the methodology perspective, XAI can be
classified into two categories: backpropagation-based and perturbation-based.
The former case, also known as gradient-based, leverages the backward pass of a
neural network to assess the influence and relevance of an input feature on the
model decision. Representative works include saliency
maps~\cite{simonyan2013deep}, Gradient class activation mapping
(Grad-CAM)~\cite{zhou2016learning}, Salient Relevance (SR)
maps~\cite{li2019beyond}, Attributes Maps (AM)~\cite{ancona2017towards},
DeepLIFT~\cite{shrikumar2017learning}, and GradSHAP~\cite{selvaraju2017grad}.
For the latter case, it primarily perturbs the input data into variants, and
then measures the change in model output to generate explanations.
Representative works include LIME~\cite{ribeiro2016should},
SHAP~\cite{lundberg2017unified}, and DeepSHAP~\cite{lundberg2017unified}. Given
that \tool\ employs Shapley value, we detail it in the following.

\noindent \textit{Shapley Value for XAI.}~The Shapley
value~\cite{shapley1953value}, initially introduced in the cooperative game
theory, quantifies each player's contribution to the total payoff of a
coalition. This concept has been adapted to machine learning to measure each
feature's contribution to a model's prediction. The Shapley value for player $i$
is determined by their average marginal contribution across all possible
coalitions, taking into account all potential player orderings. Formally, it can
be expressed as: $\phi_i(v) = \frac{1}{N} \sum_{k=1}^N
\frac{1}{\binom{N-1}{k-1}} \sum_{S\in S_k(i)} (u(S \cup \{i\}) - u(S))$ where
$N$ represents the set of all players, $S_k(i)$ is the collection of all
coalitions of size $k$ that include player $i$, $u$ is the utility function, and
$S$ is a coalition of players. In short, the Shapley value enumerates all
possible coalitions and calculates the marginal contribution of the player $i$
in each coalition. The Shapley value possesses several desirable attributes,
including efficiency, symmetry, and additivity, and uniquely satisfies the
property of locality. According to this property, a player's contribution
depends solely on the members involved in the respective coalition.

In the context of XAI, 
$N$ often represents the size of feature space, $S$ is a
subset of features, and $u$ is the model's prediction. For instance, consider
the pixel-level Shapley value for image classification. In this case, $N$ is the
number of pixels in the image, $S$ is a subset of pixels, and $u(S)$ is the
probability of the image belonging to a particular class when all pixels
excluding $S$ are masked. Likewise, in superpixel-based and concept-based
methods, $N$ is the number of superpixels or concepts, $S$ is a subset of
super-pixels or concepts, and $u(S)$ is the prediction of the model when the
remaining superpixels or concepts are masked. 


\noindent \textbf{Different Forms of Explanations.}~Given an input
image, existing XAI techniques may explain the DNN decision at the level of
individual pixels, superpixels, or concepts.

\noindent \textit{Pixel-Based XAI} explains DNN decision at the level of
individual image pixels. Prior works often use saliency
maps~\cite{simonyan2013deep, fong2017interpretable}, which highlight the most
important pixels for a given decision. Activation
maximization~\cite{mahendran2016visualizing} modifies an input image to maximize
the activation of a particular neuron or layer in a DNN to explain its decision.
Attention~\cite{gkartzonika2023learning,ntrougkas2022tame} is also employed in
XAI, offering relatively low-cost explanation in comparison to standard
gradient-based methods. However, these techniques have limitations, such as
being highly dependent on the model architecture and not providing a complete
understanding of the decision-making process. 

\noindent \textit{Superpixel-Based XAI} explains DNN decision at the level of
super-pixels, which are often perceptual groups of pixels.
SLIC~\cite{achanta2012slic} is a common superpixel algorithm that clusters
pixels together based on their spatial proximity and color similarity. SLIC
forms the basis of various mainstream XAI methods, such as Class Activation
Mapping (CAM)~\cite{zhou2016learning} and Grad-CAM~\cite{selvaraju2017grad}.
LIME offers a model agnostic approach to XAI that explains model decisions by
training a local explainable model over superpixels. RISE~\cite{petsiuk2018rise}
randomly masks parts of the input image and observes the model prediction
change. Superpixels that have the most significant impact on the output are then
used to create an importance map for explanation. 
Superpixel-based methods group pixels together to simplify the image, but this
may result in a loss of resolution and detail, particularly when attempting to
identify small or intricate details in an image. The accuracy of
superpixel-based methods is highly dependent on the quality of the image being
analyzed, and more importantly, objects that are recognized in the image.
Ill-conducted superpixel segmentation can notably undermine the accuracy of the
explanation.

\noindent \textit{Concept-Based XAI} leverages concept activation vectors
(CAVs)~\cite{kim2018interpretability} extracted from neuron activations to
distinguish between images containing user-specified concepts and random images.
It offers a quantitative way of depicting how a concept influences the DNN
decision. Furthermore, the CAV method is extended in recent works with more
flexibility and utility (e.g., with causal analysis or Shapley
values)~\cite{goyal2019explaining,ghorbani2019towards,yeh2019concept}.
However, these techniques generally require human annotation or are limited to a
pre-defined set of concepts, and this paper advocates a general and flexible
concept-based XAI technique.



\section{Method}
\label{sec:method}

\subsection{Desiderata of XAI Techniques}

XAI aims to address the opaque nature of DNNs and provide explanations for their
predictions. Aligned with prior research~\cite{li2021instance, hsiao2021roadmap,
herm2023impact}, we outline three essential desiderata for XAI techniques
applied in computer vision tasks.

\noindent \textbf{Faithfulness.}~The explanation should be highly faithful with
respect to the internal mechanism of the target DNN. Let $E$ be the explanation
and $f$ be the target DNN model under analysis, the faithfulness can be defined
as the correlation between $E$ and the actual decision-making process of $f$. A
higher correlation indicates a more effective explanation. To date, a consensus
is yet to be reached on how to quantitatively measure
faithfulness~\cite{petsiuk2018rise, samek2016evaluating, bhatt2020evaluating,
colin2022cannot}. In this paper, we employ insertion/deletion
experiments~\cite{petsiuk2018rise}, a frequently-used approach to assessing
faithfulness.

\noindent \textbf{Understandability.}~Although faithfulness is a necessary
element, it alone does not suffice for a desirable XAI process. It is equally
crucial that the explanation delivered by the XAI process is comprehensible to
humans. In other words, the XAI process must provide a lucid and
human-understandable explanation that allows users to easily grasp the reasoning
behind the model's decision-making process.

\noindent \textbf{Efficiency.}~The explanation generation process should have
relatively low computational cost so as not to cause unacceptable delays. This
ensures that the XAI technique can be practically implemented in real-world
applications without compromising system performance.

\noindent \textbf{Trade-off.}~Note that the three desiderata are often in
conflict with each other. For example, a highly faithful explanation, strictly
reflecting the internal mechanism of the target DNN, may be too obscure to
understand. Moreover, faithfulness may also impose a significant challenge to
efficiency. In fact, generating a explanation that faithfully explains the
target model is usually costly, especially for models with popular large
backbones. Hence, many existing XAI techniques trade faithfulness for efficiency
by using surrogate models to approximate the target model.
In sum, this paper advocates to strike a balance between the three desiderata to
obtain an effective XAI technique, as will be illustrated in our technical
design (\S~\ref{sec:method}) and evaluation (\S~\ref{sec:evaluation}).

\subsection{\tool\ Design}

\begin{figure}[!ht]
    \centering
    \includegraphics[width=0.7\linewidth]{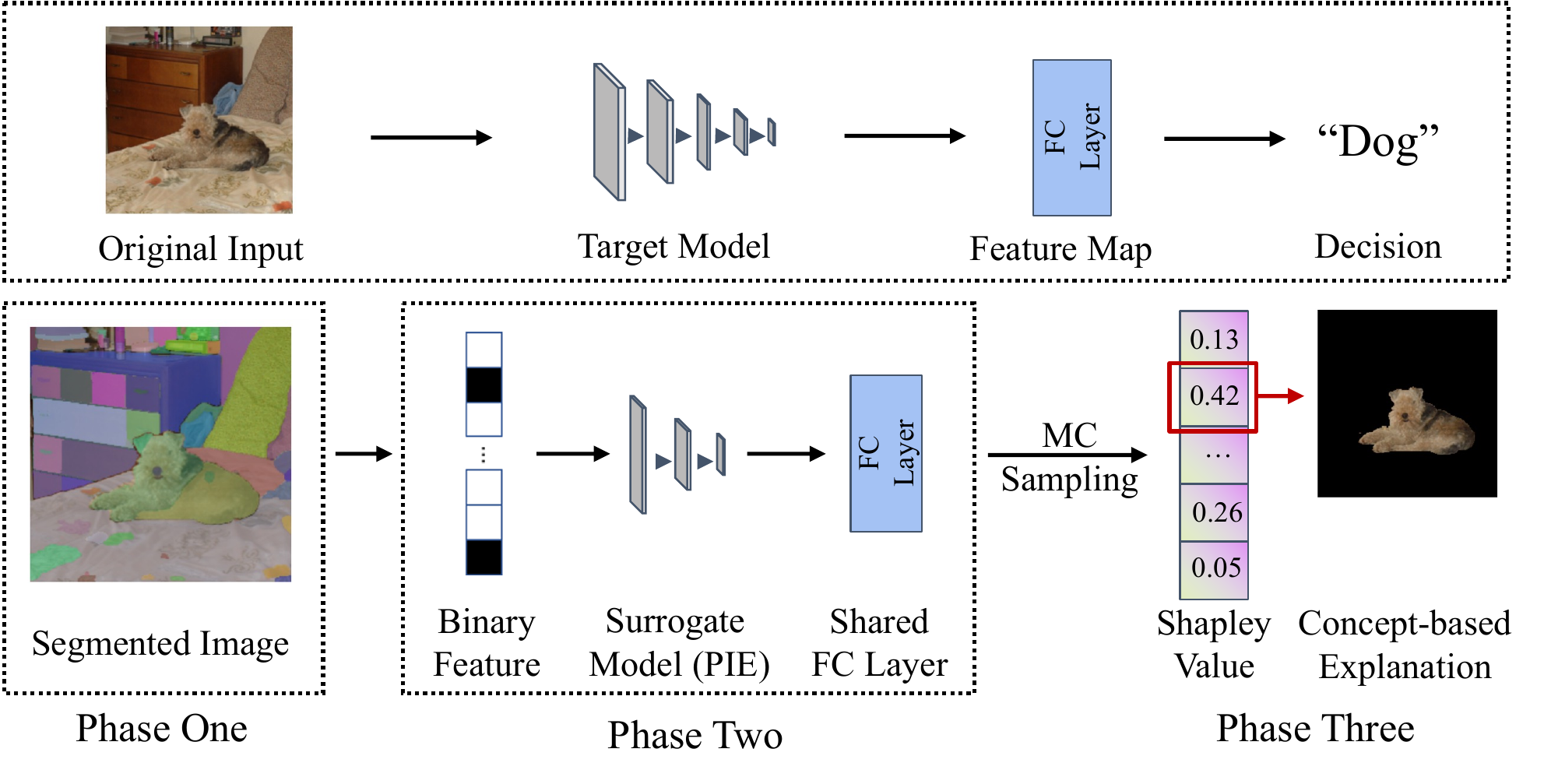}
    \caption{The technical pipeline of \tool\ in a three-phase form.}
    \label{fig:workflow}
\end{figure}

\noindent\textbf{Technical Pipeline.}~As depicted in \F~\ref{fig:workflow},
\tool\ features a three-phase pipeline to explain a DNN's prediction for an
input image. In the first phase, we employ the de facto instance segmentation
model, SAM, to partition an input image into a set of visual concepts. In the
second phase, we train a per-input equivalent (PIE) surrogate model to
approximate the behavior of the target DNN. In the third phase, we use the
surrogate model to efficiently explain the model prediction with the concepts
obtained in the first phase. In sum, \tool\ lines up SAM with XAI, such that the
de facto instance segmentation model forms the basis of XAI
\textit{faithfulness} and \textit{understandability}, whereas our novel PIE
scheme offers high \textit{efficiency} while maintaining high
\textit{faithfulness}. Below, we present the technical pipeline of \tool\ in
detail.

\noindent \textbf{Phase One: Concept Discovery.}~Concepts are defined as
prototypes that are understandable for
humans~\cite{wu2020towards,yeh2020completeness}. Traditionally, methods such as
ACE~\cite{ghorbani2019towards} and CONE-SHAP~\cite{li2021instance} leverage
superpixel method on the validation set. Then, these superpixels are clustered
into a set of concepts using its feature vectors. However, this does not
necessarily render the concepts that are understandable for humans. In this
regard, we advocate that the concepts should be semantically meaningful and
human-understandable. To this end, we employ the de facto instance segmentation
model, SAM, to obtain the concepts. 
Given an input image $x$, SAM outputs a set of instances in the image and these
instances constitute the concept for $x$. We present a sample case in ``Phase
One'' of \F~\ref{fig:workflow}. Here, we denote the set of concepts as
$\mathcal{C}=\{c_1, c_2, \cdots, c_n\}$, where $n$ is the number of concepts.

\noindent \textbf{Phase Two: Per-Input Equivalence (PIE).}~Aligned with recent
research in XAI, we use the Shapley value~\cite{shapley1953value} to identify
key concepts that contribute to the target model's prediction; the design detail
is presented below.
However, despite the general effectiveness of Shapley value, it is highly costly
to compute due to the exponential complexity. Overall, the Shapley value needs
to enumerate all possible coalitions and calculates the marginal contribution
for each concept. While the exponential complexity is usually avoided by Monte
Carlo sampling, it is still costly due to the inherent complexity of the target
model (e.g., models with large backbones).

To alleviate this hurdle, we propose the scheme of \textit{Per-Input
Equivalence} (PIE) to reduce the complexity of the target model. Intuitively, we
would expect to substitute the target model with a surrogate model $f'$ that is
computationally efficient while maintaining the same functionality of the target
model. However, it is challenging to obtain a simple model $f'$ that is fully
equivalent to the target DNN model $f$. Since each time we only explain the target
model for one certain input, we can employ a surrogate model $f'$ that is only
equivalent to $f$ over the given certain input (i.e., PIE). 
Therefore, we have $f'(\bm{b})\coloneqq f_{\text{fc}}(h(\bm{b}))$, where 
$\bm{b}$ is the one-hot encoding of the concepts in $\mathcal{C}$, $h$ mimics
the feature extractor of $f$, and $f_{\text{fc}}$ is the fully-connected (FC)
layer in $f$. 

Formally, we have $h:\{0,1\}^{|\mathcal{C}|}\to \mathbb{R}^m$ where $m$ is the
size of features in the last layer of $f$. Considering the ``Phase Two'' in
\F~\ref{fig:workflow}, where to constitute the PIE scheme, $f'$ takes the
one-hot encoding of the concepts as input, leverages $h$ to extract the feature
vector with the same semantics as $f$, and predicts the expected output of $f$
using the same FC layer (referred to as ``Shared FC Layer'' in
\F~\ref{fig:workflow}) of $f$. 
The training data for $f'$ is obtained by sampling the concepts in $\mathcal{C}$
and the corresponding probability distribution of $f$ by masking the concepts in
$\mathcal{C}$. Then, we can train $f'$ by plugging the FC layer of $f$ into $f'$
as freezed parameters and optimize $h$ with the cross-entropy loss. 

Hence, we only use the target model $f$ to obtain a few training samples for our
surrogate model $f'$ and then use $f'$ to explain $f$. Given that $f'$ is much
smaller than $f$, the PIE scheme can significantly reduce the cost of computing
the Shapley value, enabling efficient Shapley value calculations for $f$. 

\noindent \textbf{Phase Three: Concept-based Explanation.}~Given a set of visual
concepts $\mathcal{C}=\{c_1, c_2, \cdots, c_n\}$ identified in the first phase,
we aim to explain the model prediction for a given input image $x$. The
explanation can be expressed as a subset of $\mathcal{C}$, i.e., $E \subseteq
\mathcal{C}$. Recall the definition of Shapley value in \S~\ref{sec:bg}, we
consider each concept as a player and the model prediction on the original class
of $x$ as the utility function. Then, we can define the marginal contribution of
a concept $c_i$ as the difference between the model prediction on $S
\setminus\{c_i\}$ and $S$ where $S\subseteq \mathcal{C}\setminus \{c_i\}$. That
is, the marginal contribution of $c_i$ is defined as
\begin{equation}
    \Delta_{c_i}(S) = u(S\cup \{c_i\}) - u(S)
\end{equation}

Here, $u(S)\coloneqq f(\text{mask}(x, \mathcal{C}\setminus S))$ is the
prediction of the target model $f$ on the image with only concepts in $S$
(remaining concepts are masked). With the aforementioned PIE scheme, we can use
the surrogate model $f'$ to approximate $f$ (i.e., $\hat{u}(S)\coloneqq
f'(\text{mask}(x, \mathcal{C}\setminus S))$). Then, the Shapley value of $c_i$
is defined as
\begin{equation}
    \phi_{c_i}(x) = \frac{1}{n} \sum_{k=1}^n \frac{1}{\binom{n-1}{k-1}} \sum_{S\in S_k(i)} \Delta_{c_i}(S)
\end{equation}

\noindent where $S_k(i)$ is the collection of all coalitions of size $k$ that does not
contain $c_i$. Since the size of all coalitions is prohibitively large, we
approximate the Shapley value using Monte Carlo (referred to as ``MC Sampling''
in \F~\ref{fig:workflow}) sampling. In particular, we sample $K$ coalitions for
each concept and approximate the Shapley value as $\hat{\phi}_{c_i}(x) =
\frac{1}{K} \sum_{k=1}^K \Delta_{c_i}(S_k)$, where $S_k$ is the $k$-th sampled
coalition. The optimal explanation is defined as the subset of concepts that
maximizes the Shapley value, i.e.,
\begin{equation}
    E = \arg_{E\subset \mathcal{C}}\max \hat{\phi}_{E}(x)
\end{equation}

\noindent where $\hat{\phi}_{E}(x)=\sum_{c_i\in E} \hat{\phi}_{c_i}(x)$ is the Shapley
value of $E$ for $x$. Finally, we mask the concepts in $\mathcal{C}\setminus E$
and provide the masked image as the visual explanation to the user.

\noindent \textbf{Comparison with Existing Surrogate Models.}~We are aware of
existing methods such as LIME~\cite{ribeiro2016should} that also use a low cost
surrogate model to mimic the target model. We highlight the key differences
between our method and LIME. First, LIME uses a linear model as the surrogate
model, which is not expressive enough to approximate the target model. Second,
LIME learns the relation between the input and output of the target model. In
contrast, our method effectively reuses the FC layer of the target model $f$ to
form the surrogate model $f'$. This shall generally enable more accurate
approximation of the target model as the FC layer is retained.
\section{Evaluation}
\label{sec:evaluation}

In this section, we evaluate \tool from three aspects: (1) the faithfulness of
\tool in explaining the model prediction, (2) the understandability of \tool
from the human perspective, and (3) the effectiveness of PIE scheme compared
with the standard Monte Carlo sampling and surrogate model-based methods.

\subsection{\tool Faithfulness}

\noindent \textbf{Setup.}~We evaluate \tool on two popular datasets,
ImageNet~\cite{deng2009imagenet} and COCO~\cite{lin2014microsoft} and use the
standard training/validation split for both datasets. 
We use the ResNet-50~\cite{he2016deep} pre-trained on ImageNet/COCO as the
target DNN model for \tool. 

We use the \textit{insertion} and \textit{deletion} schemes to form our
evaluation metrics; these two schemes are commonly used in the literature for
evaluation of XAI techniques~\cite{petsiuk2018rise}. 
To clarify, these metrics involve generating predictions by gradually inserting
and deleting concept features from the most important to the least important
ones, and then measuring the Area Under the Curve (AUC) of prediction
probabilities. Particularly, for insertion, it starts with a fully masked image
and gradually reveals the concepts, while for deletion, it starts with a fully
unmasked image and gradually masks the concepts. Intuitively, the AUC reflects
the impact of the inserted/deleted concepts on the model prediction. For
insertion, higher AUC indicates better faithfulness, while for deletion, lower
AUC indicates better faithfulness. For both settings, we report the average
results and standard deviation of three random runs.

\noindent \textbf{Baselines.}~We compare \tool\ with nine baseline methods: (1)
DeepLIFT~\cite{shrikumar2017learning} and (2) GradSHAP~\cite{selvaraju2017grad}
are two representative backpropagation-based methods, which have been noted in
\S~\ref{sec:bg}. (3) IntGrad~\cite{sundararajan2017axiomatic}, a gradient-based
method, yields explanation by computing the path integral of all the gradients
in the straight line between an input and the corresponding reference input; (4)
KernelSHAP~\cite{lundberg2017unified} approximates Shapley values by solving a
linear regression problem; (5) FeatAbl (Feature Ablation) is a frequently-used
perturbation based approach for feature attribution~\cite{featabl}; (6)
LIME~\cite{ribeiro2016should}, a popular perturbation-based method introduced in
\S~\ref{sec:bg}, uses a surrogate model to approximate the target model. To
apply these methods, we first employ superpixel~\cite{achanta2012slic} to
generate concept patches from the input image, and then calculate their concept
importance. 

Besides, we also compare \tool\ with three variants of DeepLIFT, GradSHAP, and
IntGrad, which leverages morphological operations~\cite{yuan2022unveiling}
(e.g., image erosion and dilation~\cite{opencv_library}) as a postprocessing
step to cluster pixels into concepts. We denote these variants as DeepLIFT*,
GradSHAP*, and IntGrad*, respectively. Overall, we compare \tool\ with six
superpixel-based XAI methods and three postprocessing-based concept XAI.

\begin{table}[!htbp]
  \caption{Comparison with baseline methods across four settings. For each
  setting, the first and the second row report the mean and std.~dev.~of the
  results of three runs, respectively. $\uparrow$ and $\downarrow$ indicate
  higher and lower is better, respectively.}
  \label{tab:results}
  \centering
   \resizebox{1.00\linewidth}{!}{
  \begin{tabular}{l||c||c|c|c||c|c|c||c|c|c}
    \toprule
    & \tool & DeepLIFT & GradSHAP & IntGrad & KernelSHAP & FeatAbl & LIME & DeepLIFT*
    & GradSHAP* & IntGrad*\\
    \midrule
    \multirow{2}{*}{ImageNet/Insertion $\uparrow$} &{\bf 83.400} &75.235 &64.658 &68.772 &64.544 &70.187 &76.638 &14.707 &14.794 &15.120 \\
     &0.023&0.000 &0.035 &0.000 &0.002 &0.000 &0.027 &0.000 &0.067 &0.000 \\
    \midrule                                                                                                    
    \multirow{2}{*}{CoCo/Insertion $\uparrow$ } &{\bf 83.404} &78.199 &61.109 &65.037 &54.570 &72.260 &79.028 &8.580 &21.643 &19.755 \\
    &0.012 &0.000 &0.212 &0.000 &0.004 &0.000 &0.061 &0.000 &0.094 &0.000 \\
    \midrule       
    \multirow{2}{*}{ImageNet/Deletion $\downarrow$}  &{\bf 23.799} &25.262 &40.996 &36.214 &26.583 &37.332 &25.307 &40.620 &44.830 &46.015\\
     &0.005 &0.000 &0.061 &0.000 &0.034 &0.000 &0.064 &0.000 &0.246 &0.000\\
    \midrule                                                                                                    
                                                                                             
    \multirow{2}{*}{CoCo/Deletion $\downarrow$}  &{\bf 16.640} &17.026 &34.038 &30.074 &20.054 &26.535 &17.337 &49.697 &35.302 &38.148\\
     &0.041 &0.000 &0.144 &0.000 &0.040 &0.000 &0.049 &0.000 &0.173 &0.000\\
    \bottomrule
  \end{tabular}
 }
\end{table}

We report the effectiveness of \tool and compare it with the baseline methods in
\T~\ref{tab:results}. In particular, we observe that \tool consistently
outperforms the baseline methods across all settings. For example, in the
ImageNet dataset, \tool achieves 83.400\% AUC for insertion, which is 8.165\%
higher than the second-best method DeepLIFT. Similarly, in the COCO dataset,
\tool achieves 83.404\% AUC for insertion, which is 5.205\% higher than the
second-best method DeepLIFT. We observe similarly promising results for the
deletion evaluations.
We also observe that the standard deviation of \tool is comparable to these
baseline methods, which indicates that \tool is as stable as the majority of the
baseline methods. Moreover, it is seen that the variants of DeepLIFT, GradSHAP,
and IntGrad perform much worse than the original methods, which indicates the
incapability of those standard morphological operations in clustering pixels,
and the superiority of SAM in concept discovery.

\subsection{\tool Understandability}

To further explore the understandability of outputs from different XAI methods,
we conduct a human evaluation to assess whether \tool can generate more
human-understandable explanations than the baseline methods. To do so, we
randomly select 100 images from the ImageNet and COCO datasets, respectively. We
then generate eight explanations for each image using \tool and the seven
baseline methods (for those three morphological operation-based variants in
\T~\ref{tab:results}, we pick the best one in this evaluation). Then, for each
image, we shuffle the explanations generated by \tool and the baseline methods,
and ask our human participants (see below) to pick an explanation that they
think is most understandable.

At this step, we recruit six participants to evaluate the explanations. We
clarify that all participants are graduate students in relevant fields. We spent
about 15 minutes to train the participants to understand our task. Before
launching our study, we also provide a few sample cases to check whether the
participants understand the task. To reduce the workload, for each image and its
associated eight explanations, we randomly select three participants for the
evaluation. Thus, each participant needs to evaluate 100 explanations. We report
that each participant spends about 45 minutes to finish the evaluation.

\begin{table}[!ht]
  \caption{Human evaluation results.}
  \centering
\label{tab:human-eval}
  \resizebox{0.85\linewidth}{!}{\begin{tabular}{l||c||c|c|c||c|c|c||c||c}
\toprule
            & EAC & DeepLIFT & GradSHAP & IntGrad & KernelSHAP & FeatAbl & Lime & GradSHAP* & Discord \\
\midrule
ImageNet    & 70  & 6        & 3       & 4      & 5         & 1              & 2    & 2             & 7            \\\midrule
COCO        & 67  & 7        & 3       & 5      & 3         & 5              & 1    & 0             & 9           \\
\bottomrule
\end{tabular}
}
\end{table}

We report the results in \T~\ref{tab:human-eval}. Among 200 images in total,
participants reach a consensus (i.e., at least two out of three participants
favors the same explanation) on 184 images (92.0\%) with 93 images from ImageNet
and 91 images from COCO. Among these 184 images, \tool is favored on 137 images
(74.5\%) while the second best baseline method is only favored on seven images
(7.7\%). From \T~\ref{tab:human-eval}, it is seen that among the baseline
methods, no one is significantly better than the others.

Overall, we conclude that \tool is significantly better than the baseline
methods in terms of understandability. This is an encouraging finding that is in
line with our study in \T~\ref{tab:results}. Below, we present several cases to
demonstrate the \tool's superiority in understandability.

\begin{figure}[!ht]
  \centering
  \includegraphics[width=\linewidth]{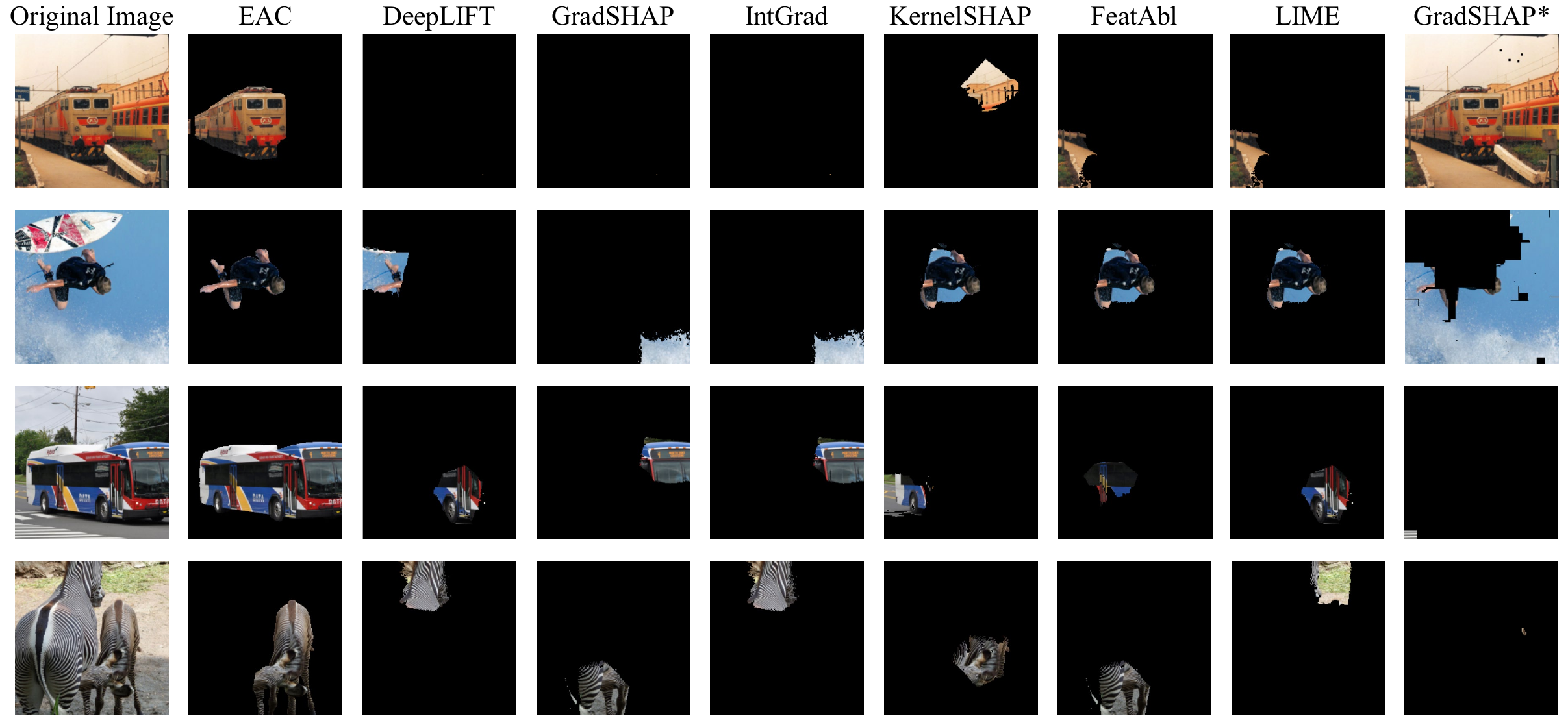}
  \caption{Sample explanations generated by \tool and the baseline methods. }
  \label{fig:case}
\end{figure}

\noindent \textbf{Case Study.}~We present several cases in \F~\ref{fig:case} to
demonstrate the effectiveness of \tool\ in explaining the model prediction.
Besides these four cases, we observe that \tool\ consistently
demonstrates its superiority in explaining the model prediction across other
test cases.
Overall, it is clear that \tool\ generates more ``well-formed,'' concept-level
explanations that are human understandable across all four cases in
\F~\ref{fig:case}. For example, in the first case, \tool\ correctly highlights
the ``train'' as the concept-level explanation, whereas the baseline methods
yield some negligible pixels (the first three baselines), a fragment of the
image (the 4th, 5th, and 6th baselines), or the entire image (the 7th baseline).
This clearly illustrates the superiority of \tool\ in terms of the faithfulness
and understandability. 

\subsection{Ablation Study of the PIE Scheme}

In this experiment, we explore the effectiveness of our proposed PIE scheme. We
compre the PIE scheme with three baselines that compute the Shapley value using
1) the original model under explanation, 2) a standard linear surrogate model
and, 3) the same model in our PIE scheme while parameter sharing is disabled.
In all settings, we use the same Monte Carlo sampling method to approximate the
Shapley value while using different schemes to represent the target DNN model. 

First, for the setting of directly using the original model, we observe a
significantly longer processing time than the others. Indeed, we report that it
takes more than 24 hours to process one image when using the same Monte Carlo
sampling method. When we slightly reduce the number of samples, the processing
time is still much longer than the other three methods (two baselines and
\tool). Similarly, we find that the AUC is also much lower than that of the
other three methods. As a result, we deem that the original model is impractical
for computing the Shapley value and a surrogate model is necessary. Accordingly,
we omit reporting the results of the original model and mark its results as
``N/A'' in \T~\ref{tab:pie}.

\begin{table}[!ht]
  \caption{Ablation study of the PIE scheme. For each setting, we report the the
    mean of the results of ten runs. Aligned with \T~\ref{tab:results},
    $\uparrow$ and $\downarrow$ indicate higher and lower is better,
    respectively.}
  \centering
\label{tab:pie}
  \resizebox{\linewidth}{!}{\begin{tabular}{c|c|c|c||c|c|c|c}
\toprule
& Model  & AUC & Processing Time (sec.) & & Model  & AUC & Processing Time (sec.) \\
\midrule
\multirow{4}{*}{ImageNet/Insertion $\uparrow$}&PIE  & 81.78  & 245 & \multirow{4}{*}{ImageNet/Deletion $\downarrow$}&PIE  & 12.47 &  244 \\
&Original Model  & N/A & N/A & &Original Model  & N/A & N/A \\
&PIE w/o PS & 50.40 & 288 & &PIE w/o PS  & 32.87 & 289 \\
&Linear Model  & 78.11 & 36 & &Linear Model  & 14.08 & 31 \\
\midrule
\multirow{4}{*}{COCO/Insertion $\uparrow$}&PIE  &87.08  & 252& \multirow{4}{*}{COCO/Deletion $\downarrow$}&PIE  & 13.71 & 203  \\
&Original Model  & N/A & N/A   & &Original Model  & N/A & N/A \\
&PIE w/o PS  & 42.86 &250 & &PIE w/o PS  & 38.36  & 222 \\
&Linear Model  & 74.86 & 67& &Linear Model  &14.36  &  131 \\
\bottomrule
\end{tabular}
}
\end{table}

We report the results of our PIE scheme and the baselines in \T~\ref{tab:pie}.
Overall, we interpret the evaluation results as highly encouraging: the PIE
scheme is notably better than all three baselines. In particular, the ablated
PIE scheme without parameter sharing (the ``PIE w/o PS'' rows in
\T~\ref{tab:pie}) is significantly worse than the PIE scheme in terms of both
AUC and the processing time. This indicates that parameter sharing is effective
in reducing the processing time while preserving high accuracy. Moreover, when
comparing with the linear surrogate model (the ``Linear Model'' rows in
\T~\ref{tab:pie}), the PIE scheme is consistently better in terms of AUC. This
indicates that the PIE scheme is more accurate, because the linear surrogate
model over-simplifies the target model. Overall, we interpret that this ablation
study illustrates the necessity of our PIE scheme, which empowers high
faithfulness and efficiency to our technical pipeline.

\section{Discussion}
\label{sec:discussion}

In this section, we first analyze the generalizability of \tool\ from the
following three aspects. 

\noindent \textbf{Different Target Models.}~In this paper, we primarily use
ResNet-50 as the target DNN model to evaluate \tool. Nevertheless, it is evident
that the technical pipeline of \tool\ is independent of particular target DNN
models. 
In short, our findings
demonstrate the persistent superiority of \tool\ over baselines when explaining
different target DNN models.

\noindent \textbf{Different Visual Domains.}~The concept extraction process for
\tool is executed using the SAM framework, which means that it functions
optimally on identical visual domains to SAM. It is advocated that SAM can be
effective on a wide range of visual domains, including medical images,
simulation images, and painting images~\cite{kirillov2023segment}. 
However, our preliminary study over some other visual domains illustrate
potentially suboptimal performance of SAM. Overall, it is clear that the
accuracy of the concept set identified by SAM significantly impacts the
performance of \tool, and ill-identified concepts may lead to suboptimal results
of \tool. We remain exploring further applications of \tool on other visual
domains for future research.

\noindent \textbf{Different Visual Tasks.}~This paper illustrates the
effectiveness of \tool on interpreting image classification, a common and core
task widely studied in previous XAI research. However, \tool is not limited to
image classification. In fact, the technical pipeline of \tool can be applied
and extended to other common visual tasks that can be explained using Shapley
value-based explanations, such as object detection.
We leave exploring \tool's potential applications on other visual tasks as
future work.

\section{Conclusion and Impact Statement}

In this paper, we propose \tool, a novel method for explaining the decision of a
DNN with any concept. \tool is based on SAM, a powerful and promotable framework
for performing precise and comprehensive instance segmentation over a given
image. We propose a highly efficient neural network pipeline to integrate the
SAM and shapley value techniques. We conduct extensive experiments to
demonstrate the effectiveness and interpretability of \tool.

\noindent \textbf{Broader Impact.}~\tool\ is a general framework for explaining
the prediction of a DNN with any concept. It can be used in many applications,
such as medical image analysis, autonomous driving, and robotics. For example,
\tool\ can be used to explain the prediction of a DNN for a medical image with
any concept, such as a tumor. This can help doctors to understand the prediction
of the DNN and improve the trustworthiness of the DNN. Overall, we believe that
\tool\ can be used to improve the trustworthiness of DNNs in many applications,
and we will publicly release and maintain \tool\ to facilitate its adoption in
the real world. 

However, in terms of \textbf{Negative Societal Impacts}, \tool\ may also be used
to explain the prediction of a DNN for a medical image with a concept that is
not related to the medical image, such as a cat, a train, or some subtle errors.
This may mislead doctors and cause serious consequences. Therefore, it may cause
harm when using \tool\ in real-world sensitive domains without sufficient safety
checks on its outputs. We will continue to improve \tool\ to reduce its negative
societal impacts. 

\bibliographystyle{unsrt}
\bibliography{bib/cv}

\end{document}